\pdfoutput=1

\documentclass[11pt]{article}

\usepackage[]{ACL2023}

\usepackage{times}
\usepackage{latexsym}

\usepackage[T1]{fontenc}

\usepackage[utf8]{inputenc}

\usepackage{microtype}

\usepackage{inconsolata}

\usepackage{graphicx}
\usepackage{booktabs}
\usepackage{amssymb}
\usepackage{lingmacros}
\usepackage{colortbl}
\usepackage{multirow}

\usepackage{CJKutf8}
\usepackage{lingmacros}
\usepackage{amsmath,amssymb,mathtools}
\usepackage{fontawesome}

\newcommand{\mathboldface}[1]{\boldsymbol{#1}}
\newcommand{\bm}[1]{\mathboldface{#1}}

\newcommand{\gl}[2]{%
\leavevmode\vtop{\hbox{#1}%
\hbox{#2\lower1.4ex\rlap{ }}}}

\title{Second Language Acquisition of Neural Language Models}

\author{Miyu Oba$^{1}$ $\;\;\;$ Tatsuki Kuribayashi$^{2,3}$ $\;\;\;$ Hiroki Ouchi$^{1,4}$  $\;\;\;$ Taro Watanabe$^{1}$ \\
 $^1$Nara Institute of Science and Technology $\;$ \\
 $^2$MBZUAI$\;$
 $^3$Tohoku University $\;$ 
 $^4$RIKEN\\
\texttt{\{oba.miyu.ol2, hiroki.ouchi, taro\}@is.naist.jp } \\
\texttt{tatsuki.kuribayashi@mbzuai.ac.ae }
}

\begin{document}
\maketitle
\begin{abstract}
With the success of neural language models (LMs), their language acquisition has gained much attention.
This work sheds light on the \textbf{second language (L2) acquisition} of LMs, while previous work has typically explored their first language (L1) acquisition.
Specifically, we trained bilingual LMs with a scenario similar to human L2 acquisition and analyzed their cross-lingual transfer from linguistic perspectives.
Our exploratory experiments demonstrated that the L1 pretraining accelerated their linguistic generalization in L2, and language transfer configurations (e.g., the L1 choice, and presence of parallel texts) substantially affected their generalizations.
These clarify their (non-)human-like L2 acquisition in particular aspects.\footnote{Our codes are available at \faGithub\ \url{https://github.com/mlieynua/sla-of-nlm}}

\end{abstract}

\section{Introduction}

Cross-lingual transferability of language models (LMs) has attracted much attention.
For example, large English LMs show some translation performance even when using a small amount of non-English languages as training data~\citep{Brown2020-zt,shi2023language}, which indicates the efficient language transfer from English to others.
Such cross-lingual transferability has been evaluated by holistic measures, such as perplexity and accuracy on downstream tasks~\citep{Papadimitriou2020-kp,Deshpande2022,Blevins2022-ta}.
On the other hand, there is much room for investigating them from \emph{linguistic perspectives}; e.g., grammatical knowledge acquisition and language transfer tendencies among languages.

\begin{figure}[t]
\centering
\includegraphics[width=1.00\linewidth]{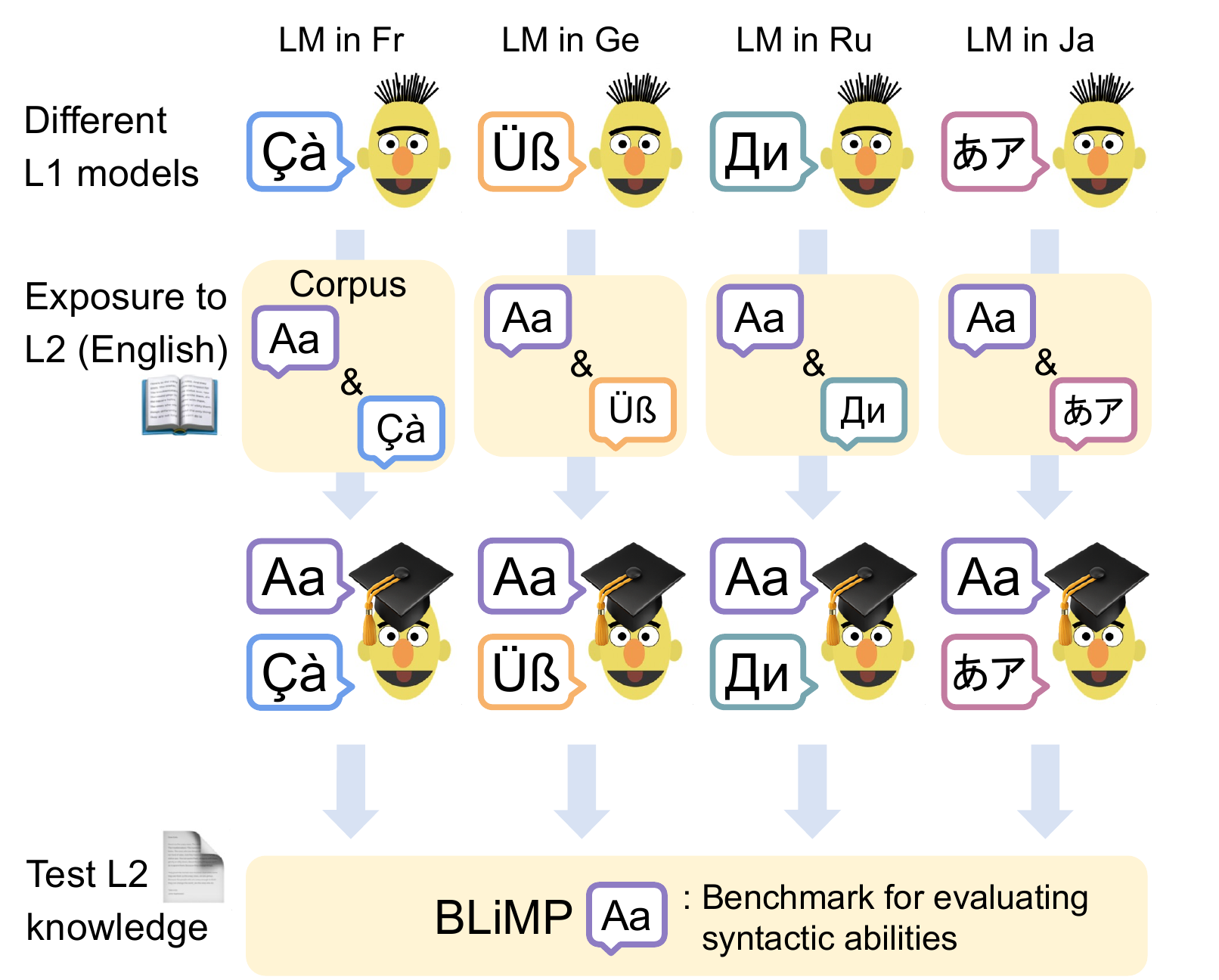}
\caption{Experimental Procedure. First, we pretrain the monolingual masked language model on the first language (first language acquisition; L1 acquisition). Then, the model is additionally trained under the bilingual setting (second language acquisition; L2 acquisition). Finally, we analyze the effect of L1 on L2 via a grammatical judgment test in L2.}
\label{fig:how_to_train_with_XLM}
\end{figure}

In this study, we investigate the cross-lingual transferability of LMs from a perspective of \textbf{second language (L2) acquisition}.
Our main research question is \emph{how first language (L1) acquisition of LMs affects the efficiency of grammar acquisition in L2}.
To answer this question, we design an experimental procedure (Section~\ref{sec:problem}):
(i) pretraining LMs in a certain language (assumed to be the L1  speakers), (ii) further training them in English as an L2, and (iii) evaluating and analyzing their linguistic generalization in L2.
As L1s, we chose four languages with different levels of difficulty in transferring to English, i.e., French, German, Russian, and Japanese.
The size of training data is restricted to match the human-like L2 acquisition scenario, which enables better comparison with human L2 acquisition tendencies and, hopefully, provides insights into L2 acquisition from a computational linguistic perspective.

We begin with exploring the inductive biases of several L2 training methods (Section~\ref{sec:pre-experiment}).
Specifically, we compared some variations of L2 data settings, such as training on only the L2 texts or on L1--L2 translation pairs.
We observed that, for example, feeding L1--L2 translation pairs into LMs slowed down their L2 grammar acquisition, compared to only feeding L2 monolingual texts every two epochs.

In our main experiments, we conducted exploratory analyses of the effects of L1 training on L2 grammar acquisition (Section~\ref{sec:experiments}).
We gained three main insights.
First, L1 knowledge promotes better linguistic generalization in L2 (Section~\ref{sec:l1-promote}).
Second, different L1s incur different generalizations in L2 (Section~\ref{sec:diff-l1}).
More specifically, Japanese and Russian are far behind French and German, which is consistent with the human-defined difficulty levels of language transfer~\citep{Chiswick2004-zz}.
Third, L1 pretraining gives different effects on different types of grammar items (Section~\ref{sec:diff-grammar}).
In particular, morphological and syntactic items get larger gains than semantic and syntax\&semantic items.

In more detail, we analyzed the \textit{process} of L2 acquisition (Section~\ref{sec:analysis}).
We investigated how L2 knowledge acquisition progresses (Section~\ref{subsec:l2_process}) and found that L2 knowledge acquisition does not progress so much until seeing the whole dataset many times (e.g., 50-100 times), implying their data inefficiency.
Furthermore, we also observed the L1 knowledge degrade during L2 acquisition; this motivates us to balance the source--target linguistic knowledge during language transfer.

\section{Second language acquisition of LMs}
\label{sec:problem}

\paragraph{Overview:}
We are interested in how L1 knowledge affects the linguistic generalization of LMs in L2.
Figure~\ref{fig:how_to_train_with_XLM} shows an overview of the experimental procedure. 
First, in our L1 acquisition simulation, we train LMs on a monolingual corpus of a specific language.
Second, in our L2 acquisition simulation, we additionally train the pretrained LMs with a corpus including L2 texts (English). 
Finally, we evaluate the grammatical judgment ability of the LMs in the L2 (English) using BLiMP~\citep{blimp}. 

\subsection{Language exposure} 
\label{subsec:exposure}

\begin{table}[t]
    \centering
    \begin{tabular}{@{}lp{1.05cm}p{0.95cm}p{1.75cm}p{0.9cm}@{}}
        \toprule
        Lang. & Family & Order & Script & Rank \\
         \cmidrule(r){1-1} \cmidrule(lr){2-2} \cmidrule(lr){3-3} \cmidrule(lr){4-4} \cmidrule(l){5-5}
        French & IE & SVO & Alphabet & $1$ \\
                \cmidrule(r){1-1} \cmidrule(lr){2-2} \cmidrule(lr){3-3} \cmidrule(lr){4-4} \cmidrule(l){5-5}
        German & IE & SOV & Alphabet & $2$  \\
                \cmidrule(r){1-1} \cmidrule(lr){2-2} \cmidrule(lr){3-3} \cmidrule(lr){4-4} \cmidrule(l){5-5}
        Russian & IE  & SVO & Cyrillic & $3$ \\
                \cmidrule(r){1-1} \cmidrule(lr){2-2} \cmidrule(lr){3-3} \cmidrule(lr){4-4} \cmidrule(l){5-5}
        Japanese & N-IE  & SOV & Kana/Kanji & $4$ \\
        \midrule
        English & IE & SVO & Alphabet & -  \\
        \bottomrule
    \end{tabular}
    \caption{Characteristics of the four languages used in our experiment. English is employed as L2, and the others are L1s. ``Rank'' indicates the transfer difficulty from the corresponding language to English, based on the linguistic distance and FSI rank; a higher value indicates a greater gap to English from the language acquisition perspective. ``Family'' indicates if Indo-European (IE) or not (N-IE). ``Order'' indicates canonical word order in the corresponding language.}
    \label{tab:langs}
\end{table}

\paragraph{First and second languages:}
We used French, German, Russian, and Japanese as L1 and employed English as L2 (Table~\ref{tab:langs}).
We expect that the transfer to English becomes more difficult in order of French, German, Russian, and Japanese from multiple perspectives: linguistic distance~\citep{grimes2002ethnologue,Chiswick2004-zz} and learning difficulty level\footnote{\url{https://www.state.gov/foreign-language-training/} Note that these difficulty levels only indicate the difficulty of transferring from English to a specific language. In our study, we tentatively assume the symmetry of the source and target language in terms of learning difficulty.}.

\paragraph{L1 acquisition:}
We first train LMs in particular L1 language using a monolingual corpus of approximately 100M words sampled from CC-100~\citep{xlmr,ccnet}. 
The corpus size is roughly similar to the number of words exposed to humans during language acquisition.
We trained the models with 100 epochs.\footnote{This number of epochs might be cognitively-implausible since humans would not face the same example 100 times, but it is also argued that the memory of language experience continues to affect the learning multiple times~\citep{Bybee2013-dc}.}

\paragraph{L2 acquisition:}
We then further train the L1 LMs under bilingual input (Section~\ref{sec:pre-experiment}).
We trained the models with 100 epochs, but the intermediate checkpoints will also be analyzed to track the process of L2 acquisition in Section~\ref{sec:analysis}.
We used Tatoeba~\citep{tatoeba}\footnote{\url{https://opus.nlpl.eu/Tatoeba.php}} as parallel corpora in L2 acquisition.
Tatoeba is a multilingual parallel corpus consisting of example sentences originally collected for human foreign language learners. 
From the L2 acquisition perspective, this amount would be large enough for human learners to learn the top 95\% English words in frequency~\citep{Nation2020-ov}.

Note that there would be several scenarios of human L2 learning/acquisition, such as through language classes or natural communications.
Following~\citet{Krashen1979-ut}, we refer to \textit{L2 acquisition} as the latter scenario of acquiring L2 through natural language exposure, e.g., raw texts.

\subsection{Learners}
We largely followed the settings of the cross-lingual language model (XLM)~\citep{conneau_xlm}, which is typically used in cross-lingual language modeling in the field of natural language processing (NLP).
In short, this is a Transformer-based bidirectional LM, but the input consists of bilingual text pairs.
The tokens in the bilingual text were randomly masked, and the model predicts the masked tokens on both L1 and L2 sides.
During the L1 training, the L1 side is the only input.

The bilingual XLM is trained from scratch (L1 training and L1--L2 training), rather than using the off-the-shelf pre-trained XLM that is trained across dozens of languages~\citep{conneau_xlm,xlmr}.
From a cognitive perspective, such a super-multilingual setting is unrealistic since humans hardly face dozens of languages in a multilingual acquisition scenario.
Rather, we hope that such a bilingual transfer will gain much attention from the adjacent areas, such as pedagogy and cognitive science of exploring human second language acquisition/learning.

Technically, we randomly initialized the parameters of the XLM (18M), constructed a bilingual tokenizer using byte pair encoding on the bilingual texts, and trained the model separately for each L1--L2 combination.
For each L1--L2 setting, we trained four models with different seeds; the results reported in Sections~\ref{sec:experiments} and~\ref{sec:analysis} are the average of scores from four models.
See Table~\ref{tab:hyperparameters} in Appendix for the hyperparameters and detailed settings.

\begin{figure}[tbp]
\centering
\includegraphics[width=\linewidth]{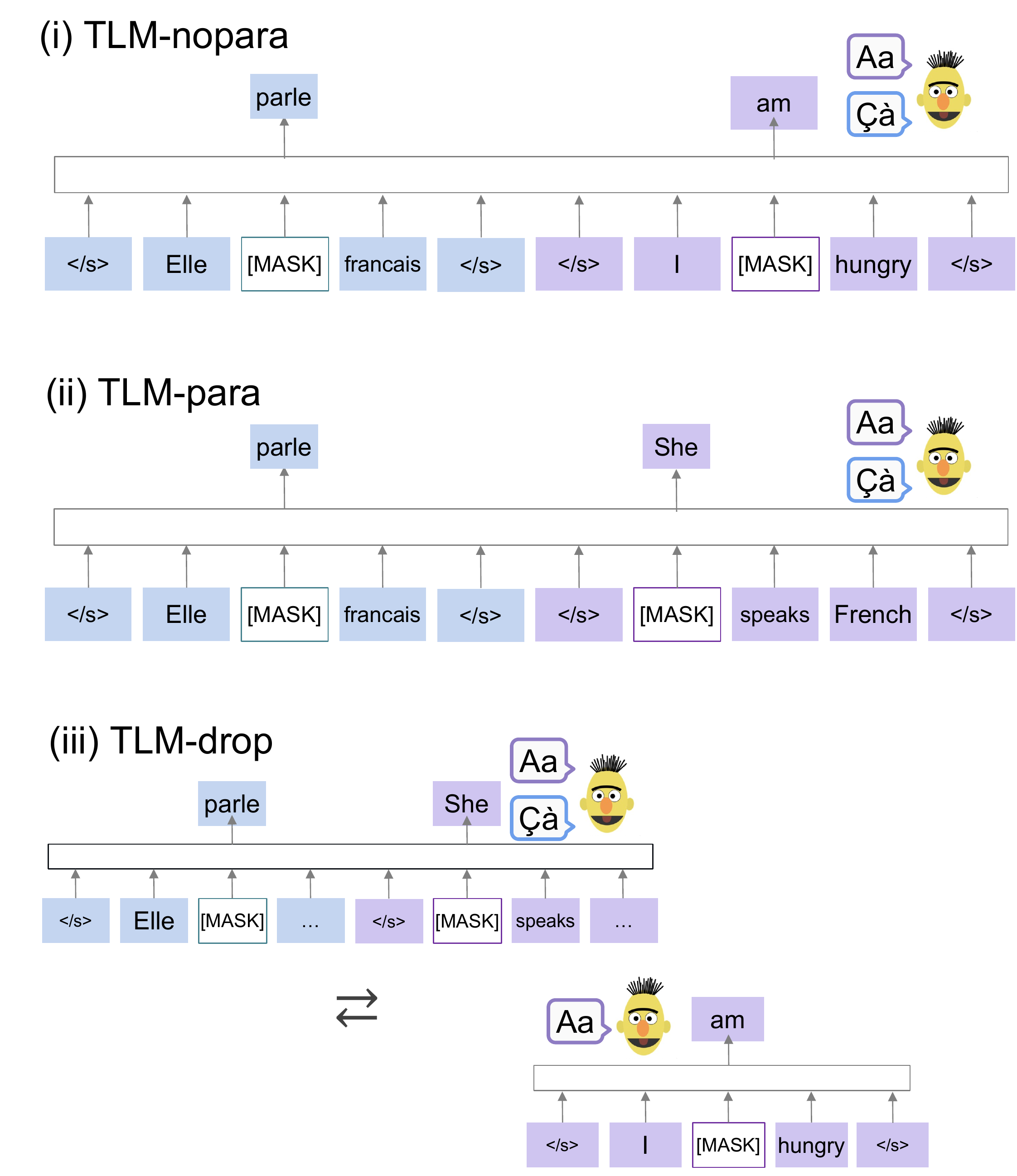}
\caption{Training settings investigated in Section~\ref{sec:pre-experiment}: (i) L1--L2 text pairs without translation relationship (TLM-nopara), (ii) translation pairs (TLM-para), and (iii) a mixed setting where parallel L1 text is removed every other epoch  (TLM-drop).}
\label{fig:conneau_model}
\end{figure}

\subsection{Evaluation}
\paragraph{Dataset:}
We used BLiMP~\citep{blimp}, a benchmark of English grammatical judgment test to evaluate the models' L2 linguistic generalization.
The dataset consists of 12 test suites; each corresponds to a specific linguistic phenomenon and falls into one of four coarse linguistic categories: morphology, syntax, semantics, and syntax\&semantics.
Each test suite has 1,000 minimal sentence pairs.
Each pair consists of grammatically acceptable and unacceptable ones as follows:

\eenumsentence{
\item Many teenagers were helping themselves.
\item * Many teenagers were helping \textbf{herself}. 
}

\paragraph{Grammatical judgement:}
To select one sentence in each pair, we adopted pseudo-perplexity, commonly used in exploring the linguistic behaviors of LMs \citep{lau-etal-2020}.
Specifically, if the model can assign a lower pseudo-perplexity to the grammatical sentence than to the paired ungrammatical one, we regard it as correct.
Following \citet{salazar-etal}, psuedo-perplexity ($\mathrm{PPPL}$) of sentence $\bm s=[w_1, w_2, \cdots, w_n]$ is computed using the bidirectional LM $\theta$:
\begin{equation}
    \mathrm{PPPL}(\bm s) = \prod_{t=1}^{n} p_{\theta}(w_t|\bm s_{\setminus w_t})^{\frac{1}{|\bm s|}} \;\;,
\end{equation}
where $w_t$ denotes the $t$-th token in sentence, and $\bm s_{\setminus w_t}$ denotes all the tokens in the sentence except for $w_t$; $[w_1, \cdots, w_{t-1}, w_{t+1}, \cdots, w_n]$.
The probability of $w_t$ given its bidirectional context $\bm s_{\setminus w_t}$ is calculated by the model $\theta$.
Based on the selected sentences, we calculated an accuracy score on each test suit of BliMP.
We also report the macro-average of accuracies among all the test suits. Note that all the accuracy scores reported in the tables/figures in this paper are multiplied by 100 for readability.

\section{Preliminary experiment: L2 exposure configurations}
\label{sec:pre-experiment}

First, we investigate the inductive bias of L2 training settings. 
While existing studies use \textit{parallel} data as an input for cross-lingual training~\citep{conneau_xlm}, we investigate the bias in this setting from L2 grammar acquisition perspectives.
\paragraph{Settings:}
We set up the three training settings with different input data: (i) L1--L2 text pairs without the translation relationship (TLM-nopara), (ii) L1-L2 translation pairs (TLM-para), and (iii) a mixed setting where L2 text concatenated with L1 parallel text is used as input or only L2 text is used as input (TLM-drop)\footnote{In other words, the bilingual and monolingual settings are switched alternately for each epoch.}.
An overview of the experimental settings is shown in Figure~\ref{fig:conneau_model} (see details in Appendix~\ref{sec:appendix}).
Note that the original XLM~\citep{conneau_xlm} adopts a setting similar to the TLM-drop.
In this experiment, we report the macro-average BLiMP accuracies across the test suites.
Table~\ref{tab:settings} shows the results (see Table~\ref{tab:full_results_l2} for the results in fine-grained test suits).

\paragraph{Translation pairs does not facilitate L2 acquisition:}
One notable point in Table~\ref{tab:settings} is that the results in the TLM-nopara setting were better than those in the TLM-para setting.\footnote{Statistical differences between the settings are tested across seeds$\times$languages with Mann-Whitney U tests ($p$=4.6e-2 for TLM-para and TLM-nopara, $p$=1.0e-2 for TLM-nopara and TLM-drop) \label{fot:stat_diff}}
This suggests that parallel data input does not facilitate L2 acquisition.
Perhaps, the TLM-para task was too easy for LMs to learn syntactic knowledge; the TLM-para task could partially be solved solely by relying on lexical knowledge, i.e., capturing the lexical correspondences between the tokens in L1 and L2 sentences and predicting the word found only in one of them.
In this sense, the TLM-nopara setting, by contrast, might impose a more difficult problem on LMs and promote their effective learning of linguistic knowledge. 

\paragraph{Switching between using L2 text with and without its parallel L1 text during training facilitates L2 acquisition:}
Another notable point is that the TLM-drop was the most effective for acquiring linguistic knowledge in L2 for LMs.\footref{fot:stat_diff}
Since we switch between using L1 text as input or not every epoch, there is a possibility that monolingual and bilingual training have a complementary positive effect.
In addition, this might mitigate the training-inference mismatch in evaluating LMs' linguistic knowledge using BLiMP.
Here, a single sentence is used as input, which is compatible with the phase of using only L2 text during the TLM-drop training.
In subsequent experiments, we will use the TLM-drop setting as it was the most effective training setting for L2 grammar acquisition.

\begin{table}[t]
    \centering
    \setlength{\tabcolsep}{2pt}
    \begin{tabular}{@{}lccrrrr}
    \toprule
        \multicolumn{1}{c}{\multirow{2}{*}{\shortstack{Model\\ (TLM)}}} &
        \multicolumn{2}{c}{Settings} &
        \multicolumn{4}{c}{First language}  \\
        \cmidrule(lr){2-3} \cmidrule(lr){4-7}
         & \textit{para.} & \textit{drop} & \multicolumn{1}{c}{Fr} & \multicolumn{1}{c}{De} & \multicolumn{1}{c}{Ru} & \multicolumn{1}{c}{Ja} \\
        \cmidrule(lr){1-1} \cmidrule(lr){2-2} \cmidrule(lr){3-3}  \cmidrule(lr){4-4}  \cmidrule(lr){5-5}  \cmidrule(lr){6-6} \cmidrule(lr){7-7} 
        nopara & & & \multicolumn{1}{r}{$52.0$} & \multicolumn{1}{r}{$57.6$} & \multicolumn{1}{r}{$51.2$} & \multicolumn{1}{r}{$52.5$} \\ 
        para & \checkmark & & \multicolumn{1}{r}{$51.1$} & \multicolumn{1}{r}{$53.6$} & \multicolumn{1}{r}{$48.9$} & \multicolumn{1}{r}{$51.3$} \\ 
        drop & \checkmark & \checkmark & \multicolumn{1}{r}{$\mathbf{58.0}$} & \multicolumn{1}{r}{$\mathbf{61.1}$} & \multicolumn{1}{r}{$\mathbf{52.8}$} & \multicolumn{1}{r}{$\mathbf{56.2}$} \\ 
        \bottomrule
    \end{tabular}
    \caption{Performance of bilingual LMs on BLiMP in different training settings. The \emph{para.} column indicates whether parallel corpus was used. The \textit{drop} column indicates whether the L1-side input is removed every other epoch.}
    \label{tab:settings}
\end{table}

\section{Experiments: L1$\rightarrow$L2 effects on linguistic generalization}
\label{sec:experiments}

\begin{table*}[t]
    \centering
    \small
    \tabcolsep 4pt
    \renewcommand{\arraystretch}{1.5}
    \begin{tabular}{@{}ccp{0.62cm}p{0.62cm}p{0.62cm}p{0.62cm}p{0.62cm}p{0.62cm}p{0.62cm}p{0.62cm}p{0.62cm}p{0.62cm}p{0.62cm}p{0.62cm}p{0.62cm}p{0.62cm}p{0.62cm}@{}}
    \toprule
        \multicolumn{3}{c}{} &
        \multicolumn{4}{c}{Morphology} &
        \multicolumn{4}{c}{Syntax} &
        \multicolumn{2}{c}{Semantics} &
        \multicolumn{2}{c}{Syntax \& Sem.}
        \\
        \cmidrule(lr){4-7} \cmidrule(lr){8-11} \cmidrule(lr){12-13} \cmidrule(lr){14-15}
        Lang. &
        L1 &
        \rotatebox{50}{\textsc{Overall}} &
        \rotatebox{50}{\textsc{Ana. agr}} &
        \rotatebox{50}{\textsc{D-n agr}} &
        \rotatebox{50}{\textsc{Irregular}} &
        \rotatebox{50}{\textsc{S-v agr}} &
        \rotatebox{50}{\textsc{Arg. str}} &
        \rotatebox{50}{\textsc{Ellipsis}} &
        \rotatebox{50}{\textsc{Filler-gap}} &
        \rotatebox{50}{\textsc{Island}} &
        \rotatebox{50}{\textsc{NPI}} &
        \rotatebox{50}{\textsc{Quantifiers}} &
        \rotatebox{50}{\textsc{Binding}} &
        \rotatebox{50}{\textsc{Ctrl. Rais.}} 
\\
        \cmidrule(r){1-1} \cmidrule(lr){2-2}  \cmidrule(lr){3-3} \cmidrule(lr){4-15} 
        \multirow{2}{*}{Fr} & \checkmark &
        \multicolumn{1}{r}{$58.0$} & 
        \multicolumn{1}{r}{$55.8$} & 
        \multicolumn{1}{r}{$69.5$} & 
        \multicolumn{1}{r}{$73.0$} & 
        \multicolumn{1}{r}{$60.4$} & 
        \multicolumn{1}{r}{$55.4$} & 
        \multicolumn{1}{r}{$67.7$} & 
        \multicolumn{1}{r}{$54.6$} & 
        \multicolumn{1}{r}{$52.2$} & 
        \multicolumn{1}{r}{$40.5$} & 
        \multicolumn{1}{r}{$56.5$} & 
        \multicolumn{1}{r}{$51.8$} & 
        \multicolumn{1}{r}{$58.6$} 
 \\
        & $\Delta$ &
        \multicolumn{1}{r}{\cellcolor[rgb]{0.409, 0.515, 0.698}{\textcolor{white}{$5.3$}}} &
        \multicolumn{1}{r}{\cellcolor[rgb]{0.513, 0.6, 0.747}{\textcolor{white}{$2.3$}}} &
        \multicolumn{1}{r}{\cellcolor[rgb]{0.088, 0.249, 0.544}{\textcolor{white}{$14.5$}}} &
        \multicolumn{1}{r}{\cellcolor[rgb]{0.461, 0.557, 0.722}{\textcolor{white}{$3.8$}}} &
        \multicolumn{1}{r}{\cellcolor[rgb]{0.287, 0.414, 0.639}{\textcolor{white}{$8.8$}}} &
        \multicolumn{1}{r}{\cellcolor[rgb]{0.341, 0.458, 0.665}{\textcolor{white}{$7.2$}}} &
        \multicolumn{1}{r}{\cellcolor[rgb]{0.0, 0.176, 0.502}{\textcolor{white}{$17.0$}}} &
        \multicolumn{1}{r}{\cellcolor[rgb]{0.435, 0.536, 0.71}{\textcolor{white}{$4.5$}}} &
        \multicolumn{1}{r}{\cellcolor[rgb]{0.619, 0.688, 0.798}$-0.8$} &
        \multicolumn{1}{r}{\cellcolor[rgb]{0.471, 0.565, 0.727}{\textcolor{white}{$3.5$}}} &
        \multicolumn{1}{r}{\cellcolor[rgb]{0.633, 0.699, 0.804}$-1.2$} &
        \multicolumn{1}{r}{\cellcolor[rgb]{0.525, 0.611, 0.753}{\textcolor{white}{$1.9$}}} &
        \multicolumn{1}{r}{\cellcolor[rgb]{0.539, 0.622, 0.76}{\textcolor{white}{$1.5$}}} 
        
\\
        \cmidrule(r){1-1} \cmidrule(lr){2-2}  \cmidrule(lr){3-3} \cmidrule(lr){4-15} 
        \multirow{2}{*}{De} & \checkmark &
        \multicolumn{1}{r}{$61.1$} &
        \multicolumn{1}{r}{$43.1$} & 
        \multicolumn{1}{r}{$68.7$} & 
        \multicolumn{1}{r}{$69.3$} & 
        \multicolumn{1}{r}{$67.0$} & 
        \multicolumn{1}{r}{$53.1$} & 
        \multicolumn{1}{r}{$63.5$} & 
        \multicolumn{1}{r}{$68.2$} & 
        \multicolumn{1}{r}{$47.7$} & 
        \multicolumn{1}{r}{$54.6$} & 
        \multicolumn{1}{r}{$80.5$} & 
        \multicolumn{1}{r}{$65.2$} & 
        \multicolumn{1}{r}{$52.2$} 
 \\
        & $\Delta$ & 
        \multicolumn{1}{r}{\cellcolor[rgb]{0.412, 0.517, 0.699}{\textcolor{white}{$5.2$}}} &
        \multicolumn{1}{r}{\cellcolor[rgb]{0.385, 0.495, 0.686}{\textcolor{white}{$5.9$}}} &
        \multicolumn{1}{r}{\cellcolor[rgb]{0.206, 0.347, 0.601}{\textcolor{white}{$11.1$}}} &
        \multicolumn{1}{r}{\cellcolor[rgb]{0.992, 0.996, 0.976}$-11.5$} &
        \multicolumn{1}{r}{\cellcolor[rgb]{0.093, 0.253, 0.546}{\textcolor{white}{$14.3$}}} &
        \multicolumn{1}{r}{\cellcolor[rgb]{0.431, 0.532, 0.708}{\textcolor{white}{$4.6$}}} &
        \multicolumn{1}{r}{\cellcolor[rgb]{0.08, 0.243, 0.54}{\textcolor{white}{$14.7$}}} &
        \multicolumn{1}{r}{\cellcolor[rgb]{0.425, 0.528, 0.705}{\textcolor{white}{$4.8$}}} &
        \multicolumn{1}{r}{\cellcolor[rgb]{0.435, 0.536, 0.71}{\textcolor{white}{$4.5$}}} &
        \multicolumn{1}{r}{\cellcolor[rgb]{0.251, 0.384, 0.622}{\textcolor{white}{$9.8$}}} &
        \multicolumn{1}{r}{\cellcolor[rgb]{0.432, 0.533, 0.708}{\textcolor{white}{$4.6$}}} &
        \multicolumn{1}{r}{\cellcolor[rgb]{0.545, 0.627, 0.763}{\textcolor{white}{$1.4$}}} &
        \multicolumn{1}{r}{\cellcolor[rgb]{0.67, 0.73, 0.822}$-2.2$}
        
\\
        \cmidrule(r){1-1} \cmidrule(lr){2-2}  \cmidrule(lr){3-3} \cmidrule(lr){4-15} 
        \multirow{2}{*}{Ru} & \checkmark & 
        \multicolumn{1}{r}{$52.8$} & 
        \multicolumn{1}{r}{$52.9$} & 
        \multicolumn{1}{r}{$58.6$} & 
        \multicolumn{1}{r}{$72.7$} & 
        \multicolumn{1}{r}{$54.9$} & 
        \multicolumn{1}{r}{$47.0$} & 
        \multicolumn{1}{r}{$54.2$} & 
        \multicolumn{1}{r}{$52.4$} & 
        \multicolumn{1}{r}{$49.3$} & 
        \multicolumn{1}{r}{$32.8$} & 
        \multicolumn{1}{r}{$56.2$} & 
        \multicolumn{1}{r}{$40.7$} & 
        \multicolumn{1}{r}{$61.4$} 
 \\
        & $\Delta$ & 
        \multicolumn{1}{r}{\cellcolor[rgb]{0.569, 0.646, 0.774}{\textcolor{white}{$0.7$}}} &
        \multicolumn{1}{r}{\cellcolor[rgb]{0.699, 0.754, 0.836}$-3.1$} &
        \multicolumn{1}{r}{\cellcolor[rgb]{0.48, 0.573, 0.732}{\textcolor{white}{$3.2$}}} &
        \multicolumn{1}{r}{\cellcolor[rgb]{0.581, 0.656, 0.78}$0.3$} &
        \multicolumn{1}{r}{\cellcolor[rgb]{0.491, 0.582, 0.737}{\textcolor{white}{$2.9$}}} &
        \multicolumn{1}{r}{\cellcolor[rgb]{0.576, 0.652, 0.778}$0.5$} & \multicolumn{1}{r}{\cellcolor[rgb]{0.412, 0.516, 0.699}{\textcolor{white}{$5.2$}}} &
        \multicolumn{1}{r}{\cellcolor[rgb]{0.451, 0.549, 0.718}{\textcolor{white}{$4.1$}}} &
        \multicolumn{1}{r}{\cellcolor[rgb]{0.66, 0.721, 0.817}$-1.9$} &
        \multicolumn{1}{r}{\cellcolor[rgb]{0.554, 0.634, 0.767}{\textcolor{white}{$1.1$}}} &
        \multicolumn{1}{r}{\cellcolor[rgb]{0.747, 0.794, 0.859}$-4.5$} &
        \multicolumn{1}{r}{\cellcolor[rgb]{0.595, 0.668, 0.787}$-0.1$} &
        \multicolumn{1}{r}{\cellcolor[rgb]{0.581, 0.656, 0.78}$0.3$} 
        
\\
        \cmidrule(r){1-1} \cmidrule(lr){2-2}  \cmidrule(lr){3-3} \cmidrule(lr){4-15} 
        \multirow{2}{*}{Ja} & \checkmark & 
        \multicolumn{1}{r}{$56.2$} & 
        \multicolumn{1}{r}{$61.5$} & 
        \multicolumn{1}{r}{$65.8$} & 
        \multicolumn{1}{r}{$70.5$} & 
        \multicolumn{1}{r}{$53.0$} & 
        \multicolumn{1}{r}{$52.1$} & 
        \multicolumn{1}{r}{$55.3$} & 
        \multicolumn{1}{r}{$51.3$} & 
        \multicolumn{1}{r}{$54.0$} & 
        \multicolumn{1}{r}{$41.0$} & 
        \multicolumn{1}{r}{$50.6$} & 
        \multicolumn{1}{r}{$61.0$} & 
        \multicolumn{1}{r}{$57.8$} 
 \\
        & $\Delta$ & 
        \multicolumn{1}{r}{\cellcolor[rgb]{0.54, 0.623, 0.76}{\textcolor{white}{$1.5$}}} &
        \multicolumn{1}{r}{\cellcolor[rgb]{0.568, 0.646, 0.774}{\textcolor{white}{$0.7$}}} &
        \multicolumn{1}{r}{\cellcolor[rgb]{0.419, 0.522, 0.702}{\textcolor{white}{$5.0$}}} &
        \multicolumn{1}{r}{\cellcolor[rgb]{0.708, 0.761, 0.84}$-3.3$} &
        \multicolumn{1}{r}{\cellcolor[rgb]{0.553, 0.634, 0.767}{\textcolor{white}{$1.1$}}} &
        \multicolumn{1}{r}{\cellcolor[rgb]{0.524, 0.609, 0.752}{\textcolor{white}{$2.0$}}} &
        \multicolumn{1}{r}{\cellcolor[rgb]{0.441, 0.54, 0.713}{\textcolor{white}{$4.3$}}} &
        \multicolumn{1}{r}{\cellcolor[rgb]{0.553, 0.634, 0.767}{\textcolor{white}{$1.1$}}} &
        \multicolumn{1}{r}{\cellcolor[rgb]{0.432, 0.533, 0.708}{\textcolor{white}{$4.6$}}} &
        \multicolumn{1}{r}{\cellcolor[rgb]{0.515, 0.602, 0.748}{\textcolor{white}{$2.2$}}} &
        \multicolumn{1}{r}{\cellcolor[rgb]{0.692, 0.748, 0.833}$-2.9$} &
        \multicolumn{1}{r}{\cellcolor[rgb]{0.565, 0.643, 0.772}{\textcolor{white}{$0.8$}}} &
        \multicolumn{1}{r}{\cellcolor[rgb]{0.514, 0.601, 0.748}{\textcolor{white}{$2.3$}}}
        
 \\
        \bottomrule
    \end{tabular}
    \caption{English (L2) grammatical knowledge of bilingual LMs with different L1s. \textsc{Overall} indicates the macro-average accuracy over all linguistic phenomena. The rows with \checkmark in the L1 column exhibit the accuracy of bilingual LMs on BLiMP. The rows with $\Delta$ in the L1 column exhibit the performance difference between the LMs with and without L1 pretraining. The coarse categories (e.g., Morphology) are from the metadata of BLiMP. 
    }
    \label{tab:pretrain}
\end{table*}

\begin{table}[t]
    \centering
    \small
    \begin{tabular}{lrrrr}
        \toprule
         L1 & Morph. & Syntax & Semantics & Syn.\&Sem. \\
         \cmidrule(r){1-1} \cmidrule(lr){2-2} \cmidrule(lr){3-3} \cmidrule(lr){4-4} \cmidrule(lr){5-5}
        Fr & \multicolumn{1}{r}{$7.3$} & \multicolumn{1}{r}{$7.0$} & \multicolumn{1}{r}{$1.2$} & \multicolumn{1}{r}{$1.7$} \\
        De & \multicolumn{1}{r}{$5.0$} & \multicolumn{1}{r}{$7.2$} & \multicolumn{1}{r}{$7.2$} & \multicolumn{1}{r}{$-0.4$} \\
        Ru & \multicolumn{1}{r}{$0.8$} & \multicolumn{1}{r}{$1.9$} & \multicolumn{1}{r}{$-1.7$} & \multicolumn{1}{r}{$0.1$} \\
        Ja & \multicolumn{1}{r}{$0.9$} & \multicolumn{1}{r}{$3.0$} & \multicolumn{1}{r}{$-0.3$} & \multicolumn{1}{r}{$1.5$} \\
                \cmidrule(r){1-5}
         Avg. & \multicolumn{1}{r}{$\mathbf{3.5}$} & \multicolumn{1}{r}{$\mathbf{4.8}$} & \multicolumn{1}{r}{$\mathbf{1.6}$} & \multicolumn{1}{r}{$\mathbf{0.7}$} \\
        \bottomrule
    \end{tabular}
    \caption{Performance difference between the LMs with and without L1 pretraining for each coarse category of grammatical items (morphology, syntax, semantics, and syntax\&semantics). 
    }
    \label{tab:category}
\end{table}

\begin{figure*}[t]
\centering
\includegraphics[width=1.00\linewidth]{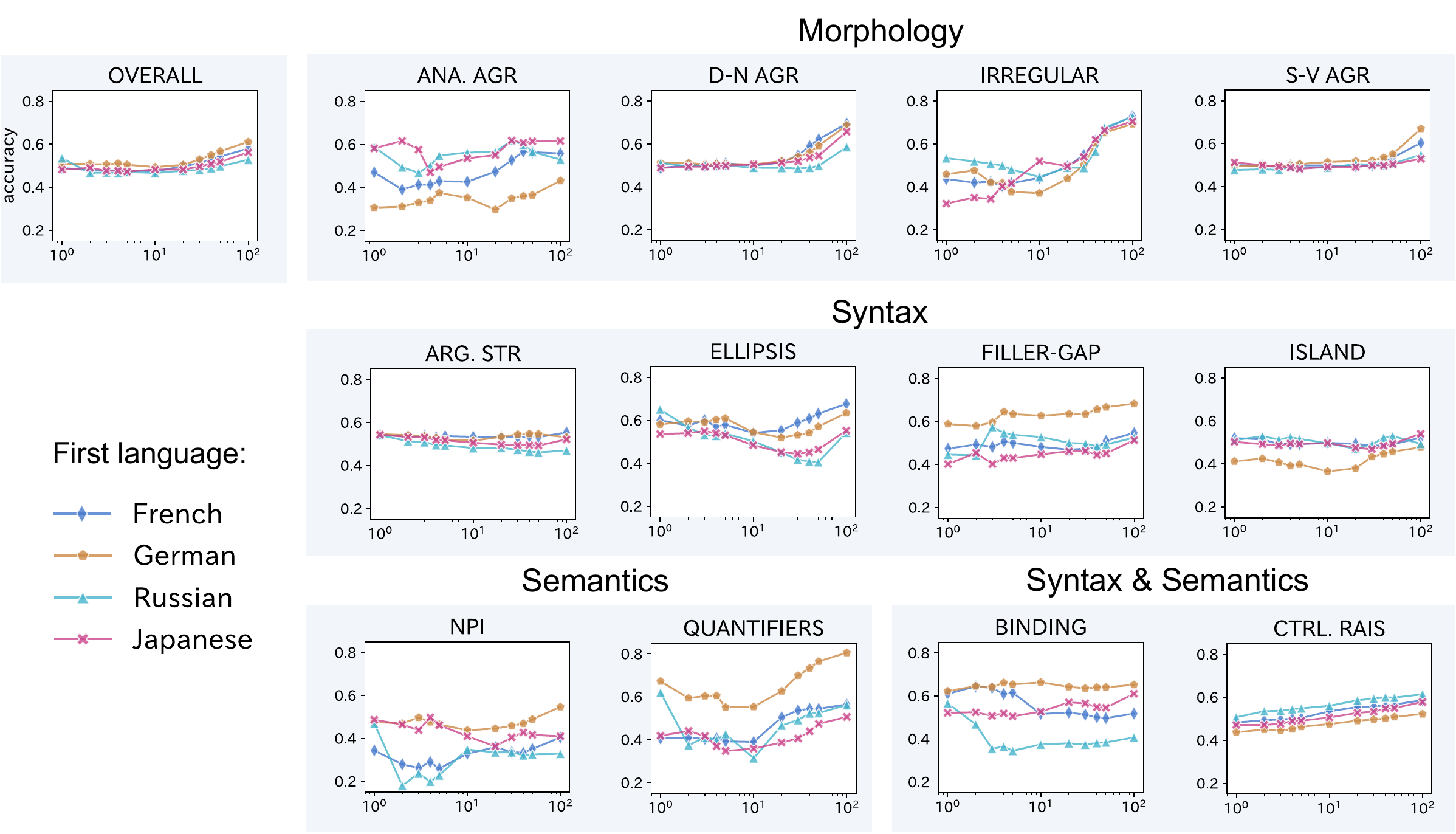}
\caption{Grammar learning trajectories in each test suite on BLiMP (the L2 side). The x-axis denotes the epoch during L2 acquisition, and the y-axis denotes the accuracy in the corresponding test suite.}
\label{fig:l1_all_trajectories}
\end{figure*}

We investigate how \emph{pretraining with L1} affects L2 grammar acquisition in LMs.
We exploratorily compare the linguistic generalization of LM trained in the settings with or without L1 pretraining.
Table~\ref{tab:pretrain} shows the grammaticality judgment ability after additional training.
The \textsc{Overall} column indicates the macro-average accuracy score across the grammar items.
The $\Delta$ rows show the difference in BLiMP accuracy between models with and without pretraining.
Here, the models without pretraining were trained only with bilingual corpus without L1 monolingual corpus pretraining.

\subsection{L1s promote L2 generalization}
\label{sec:l1-promote}
Table~\ref{tab:pretrain} shows the effect of pretraining with L1 on L2 grammar acquisition.
Most of the $\Delta$ values are positive; i.e., the models pretrained with L1s achieved better results than those without pretraining.
This demonstrates that pretraining in a particular language generally improves English grammatical ability.\footnote{The standard deviation over the different seeds is an average of 1.2 (0–100 scale) for the scores in Table~\ref{tab:pretrain}, which means that the seed randomness does not overturn the conclusion.}
This positive effect is in light of the assumptions that different languages share some grammatical universals, and learners could use such a common property in language transfer~\cite{cook1985chomsky}.
\citet{ri-language-transfer} also reported that pretraining in a natural language other than English improves the overall syntactic parsing performance in English.
Our results are consistent with such positive effects.
Besides, our experiment provides the results of each of more fine-grained grammatical items related to morphological, syntactic, and semantic phenomena.

\subsection{Differences in L1s}
\label{sec:diff-l1}
The $\Delta$ values in the \textsc{Overall} column in Table~\ref{tab:pretrain} differ across the L1s.
French is the highest, followed closely by German, and Japanese and Russian are far behind these two languages; pretraining in French and German is much more effective than in Japanese and Russian.
This ordering shows parallels with the presumed language learning difficulty order: French, German, Russian, and Japanese.
This suggests that the difficulty of acquiring an L2 grammatical ability is somewhat close between LMs and humans.

\subsection{Differences in grammar items}
\label{sec:diff-grammar}
The $\Delta$ scores in Table~\ref{tab:pretrain} exhibit that different grammatical items obtain different degrees of gains.
Table~\ref{tab:category} shows the average $\Delta$ scores for each course grammar category.
There was a general tendency for morphological and syntactic items to get larger gains from the L1 pretraining than semantic and syntax\&semantic items except for particular settings, e.g., \textsc{Irregular}.
It has been shown that linguistic phenomena related to semantics such as \textsc{NPI} (negative polarity item) and \textsc{Quantifiers} were relatively difficult  for LMs to learn~\cite{blimp}.
Based on this, there is a concern that LMs failed to learn enough such linguistic knowledge in L1 to transfer it to another language.

\subsection{Differences in L1$\times$grammar-item}
Notably, in specific combinations of L1s and grammar items, the L1 pretraining hurt the L2 generalization, i.e., negative transfer problem.
For example, the performance in the \textsc{Irregular} item was not so much enhanced or even degraded by L1 pretraining.
The \textsc{Irregular} (Irregular forms) item targets English-specific irregular verb conjugations; its less effect by L1 pretraining is due to the concern that the \textit{irregular} patterns generally could not be predictable by other language's knowledge.

We also found that the same grammatical item was affected differently depending on the L1.
For example, the $\Delta$ values on the \textsc{Filler-gap} item in Table~\ref{tab:pretrain} differ across the L1s, e.g., 4.8 in German and 1.1 in Japanese.
At least in this \textsc{Filler-gap} aspect, there is an interesting parallel between our results and linguistic notions; Japanese is the only language where gap precedes filler in wh-construction among the L1s we used, and the transfer from Japanese to English was indeed limited ($\Delta=1.1$) compared to other L1s.
There might be a possibility that such linguistic (dis)similarities are reflected in the results.
Nevertheless, concluding the exact consistency between our L1$\times$grammar-item results (Table~\ref{tab:pretrain}) and the L1--L2 grammatical similarity requires further interdisciplinary research.

\section{Analysis: acquisition process}
\label{sec:analysis}
This section sheds light on the \textit{process} of L2 acquisition.
We investigate how L2 knowledge acquisition progresses (Section~\ref{subsec:l2_process}) and how original L1 knowledge changes during L2 acquisition (Section~\ref{subsec:l1-knowledge}).
As for the L1 knowledge during L2 acquisition, there is a concern, for example, that LMs exhibit catastrophic forgetting about their L1.

\begin{figure*}[t]
\centering
\includegraphics[width=1.00\linewidth]{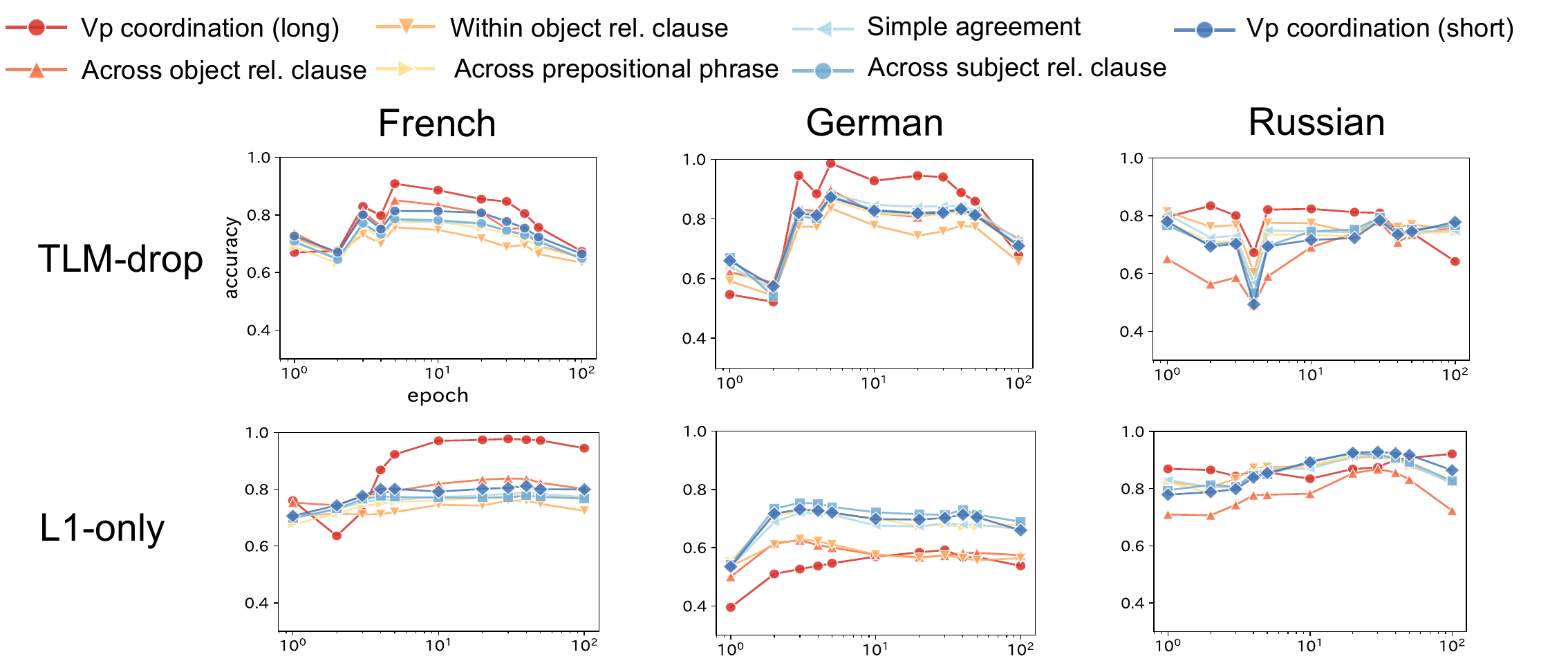}
\caption{Grammar learning trajectories in each test suite on CLAM (the L1 side). The x-axis denotes the epoch during L2 acquisition, and the y-axis denotes the accuracy in the corresponding test suite. The upper parts show the scores of the bilingual LMs (L1 pretraining and bilingual training). The lower parts show the scores of the L1-only LMs (L1 pretraining and further training on L1 texts collected from the parallel corpus).}
\label{fig:l1_trajectories}
\end{figure*}

\subsection{L1$\rightarrow$L2 effects}
\label{subsec:l2_process}

\paragraph{Settings: }
We evaluate the L2 linguistic knowledge of intermediate checkpoints of LMs (Figure~\ref{fig:l1_all_trajectories}).
Specifically, we evaluate LMs after \{1, 2, 3, 4, 5, 10, 20, 30, 40, 50, 100\} training epochs.
Note that we analyzed the same models used in Section~\ref{sec:experiments}; the results after 100 epochs are the same ones reported in Section~\ref{sec:experiments}.

\paragraph{General improvement after dozens of epochs: }
The trajectory of the \textsc{Overall} scores in Figure~\ref{fig:l1_all_trajectories}  suggest that linguistic ability generally improves along with the number of epochs.
There was a tendency for large improvements to emerge after dozens of epochs; in other words, the models began to acquire L2 knowledge after seeing the same examples many times, e.g., 50-100 times.
Note that humans are argued to acquire a vocabulary after encountering the same word about 12 times~\cite{Nation2020-ov}, and of course, the lexical and syntactic acquisition is not comparable, but the observation that the L2 knowledge improves after 50-100 rounds of the corpus may be in the direction that LMs are inefficient at acquiring a new language.

\paragraph{Differences in grammar items: }
Focusing on the general trajectory shapes for each grammatical item, we observed at least four patterns: (i) spike-at-the-end (\textsc{D-N agr.}, \textsc{Irregular}, \textsc{S-V agr.}), (ii) flat (\textsc{Arg.str.}, \textsc{Ctrl.Rais.}, \textsc{Island}), (iii) bumpy (\textsc{Ana.agr.}, \textsc{Ellipsis}, \textsc{NPI}, \textsc{Quantifiers}), and (iv) mixed (\textsc{Filler-gap}, \textsc{Binding}).
In addition, these groups roughly mirror the linguistic categories of the grammar items (morphology, syntax, semantics, and syntax\&semantics); for example, all the items in the spike-at-the-end group are morphological phenomena, while all the semantic categories (\textsc{NPI}, \textsc{Quantifiers}) yielded the bumpy patterns.
Note that existing studies reported that low-level (e.g., morphological) linguistic skills could be acquired earlier and vice versa~\citep{liu2021probing,Blevins2022-ta}; but at least in our cognitively-inspired bilingual training scenario, we did not observe such an explicit tendency.

\paragraph{Inter-L1s differences: }
In terms of the change of inter-L1s differences of accuracies in each grammar item, there are several different patterns: (i) converging (\textsc{Irregular}, \textsc{Island}), (ii) diverging (\textsc{Arg.str.}, \textsc{Binding}, \textsc{D-N agr.}, \textsc{S-V agr.}), and (iii) none of them.
Considering \textsc{Irregular} as an example of the converging group, the accuracies were substantially different across the L1s in the initial stage of training; these differences, however, gradually reduced along with epochs.
On the other hand, considering \textsc{S-V agr.} as an example of the diverging group, the accuracies gradually differed in the latter stage of training among the LMs.
In the third group, the inter-L1s accuracy differences remain the same or unstable during L2 tranining (\textsc{Ana.agr.}, \textsc{Ctrl.Rais.}, \textsc{Ellipsis}, \textsc{Filler-gap}, \textsc{NPI}, \textsc{Quantifiers}).
At least the former two groups imply that pretraining with different L1s differently affects the efficiency of L2 acquisition (e.g., different slopes for different L1s).

\vspace{\baselineskip}
To sum up, we clarified that different L1s and grammar items exhibit different learning dynamics of LMs.
The cognitive plausibility of these patterns could be the next important investigation.

\subsection{L2$\rightarrow$L1 effects}
\label{subsec:l1-knowledge}
In contrast to the previous analysis, which analyzed the impact in the L1$\rightarrow$L2 direction, we further analyze the L2$\rightarrow$L1 impact.
In applied linguistics, L1--L2 impact in both directions is of interest; for example, it is reported that the L1 ability is sometimes hurt by the increase of L2 exposure~\citep{kecskes2008effect,Haman2017-gm}.

\paragraph{Settings: }
In the same way as Section~\ref{subsec:l2_process}, we evaluate the grammatical knowledge of LMs during L2 training, but the focus is on the \textbf{L1} knowledge.
We used a multilingual benchmark of grammatical judgment test CLAMS~\citep{mueller-etal}.
This dataset consists of seven syntactic test suites for several languages.
Similarly to BLiMP, the dataset consists of pairs of sentences, where one is grammatical, and the other is ungrammatical.
We report the accuracy scores in terms of whether the LMs could assign lower pseudo-perplexity to the grammatically correct sentence.
We analyze French-L1, German-L1, and Russian-L1 LMs since this dataset covers these three L1s.

As a baseline, we also evaluate the L1-only LMs that are first pretrained in L1, then additionally trained with the L1-side texts collected from the corresponding parallel corpus; that is, the only difference between bilingual LMs and L1-only baselines was the presence of L2 texts during the L2 acquisition phase.

\paragraph{L2 effects once occur, but diminish: }
Figure~\ref{fig:l1_trajectories} shows the results (see Table~\ref{tab:full_results_l1} for the exact scores).
We found that the L1 knowledge is greatly influenced by the L2, especially at the initial stage of L2 training.
For example, the French-L1 and German-L1 LMs gained positive effects, and the Russian-L1 LM got a negative influence (the top row of Figure~\ref{fig:l1_trajectories}).
In addition, in the latter stage, the L2 effects on the L1 knowledge gradually fade, in either a good or bad sense, and the L1 linguistic knowledge is converging to the original level.
For example, French-L1 and German-L1 LMs exhibited better performance after bilingual language modeling once, e.g., at the 10 epochs, but the gain decreased after further bilingual training.

\paragraph{L2 negatively affects L1 knowledge: }
In Figure~\ref{fig:l1_trajectories}), the L1 knowledge in bilingual LMs was competitive or even inferior to that in L1-only LMs in the end, although there were also exceptional cases in specific grammar items in German.
For example, in the French-L1 LMs, the L1 syntactic generalization performance after L2 training converged below 0.7 points, while the L1-only baseline model generally achieved above 0.7 points.
Contrasting the L1 results with L2 ones (Section~\ref{subsec:l2_process}), there is an asymmetry effect between L1 and L2.
From a perspective of developing linguistically better multi-lingual LMs, these suggest that balancing the linguistic knowledge of the languages, especially enhancing the L1 knowledge, during language transfer is challenging even if the model is exposed to L1 during bilingual training.
Addressing these challenges with, for example, some regularization will be a promising direction from an engineering perspective.

\section{Related work}

\paragraph{Cognitively-motivated analysis of neural LMs: }
Investigations into the ability of neural models in language acquisition began in the 1980s with the interest of whether language can be acquired without innate knowledge~\citep{Rumelhart1985-xj,Pinker1988-zb}.
The initial investigation was made with simple neural networks; after the development of Neural NLP~\citep{Manning2015-mm}, the classical questions posed by cognitive science are currently revisited~\citep{Kirov2018-yf}.
A typical movement is a growing interest in the \textit{probing} of neural LMs' linguistic knowledge~\citep{lizen-2016, Warstadt-BERT}. 
In this context, our study analyzes L2 acquisition in neural LMs, while existing studies have typically focused on L1 acquisition.

\paragraph{Language transfer in computational models: }
Language transfer of NLP models is actively researched from both engineering and scientific perspectives.
In the engineering context, to mitigate the English-centric focus of NLP techniques, models that can handle more languages have been developed~\citep{Dong2015,conneau_xlm,xlmr}.
From the scientific perspective, the mechanism and linguistic properties of LMs' language transfer have been explored~\citep{noauthor_undated-ut,Chang2022-hb,Blevins2022-ta}, sometimes beyond the transfer between natural languages~\citep{ri-language-transfer,Papadimitriou2020-kp}.
One of the motivations of such analyses is to quantify the transferable universals behind (non-)languages.
Notably, simulation of L2 acquisition is also explored from pedagogical motivations~\citep{settles-etal-2018}.

\paragraph{Language transfer in humans: }
L2 acquisition/learning has long been studied in applied linguistics, psycholinguistics, and pedagogy fields~\citep{krashen1981second,hatch1983psycholinguistics,ellis2010second}.
These fields articulated several hypotheses/theories on human language learning, e.g., input hypothesis~\citep{krashen1977some}.
Analyzing the LMs' L2 acquisition in a more direct light with these hypotheses would be interesting for future work.

\section{Conclusions}
We have investigated the L2 acquisition of LMs, especially focusing on their grammatical knowledge in L1 and L2.
Specifically, we have trained bilingual LMs under a similar scenario to the human L2 acquisition and then analyzed their cross-lingual transfer.
Our experiments have demonstrated that L1 pretraining promotes their linguistic generalization in L2, and there are interesting variations in L1 pretraining effects with respect to the L1 choice, training settings, and grammar items.
The results have also implied that their L2 acquisition is not human-like in particular aspects.

\section*{Limitations}

\subsection*{Coverage of experiments}
There is room for further exploration in terms of model architectures, data, and evaluation settings in our study.
Experiments in more diverse settings would enhance the generality of the conclusions.

\paragraph{Models:} The architecture was fixed to XLM (14M), although we tested four models with different seeds in each setting.
Specifically, testing unidirectional LMs will be in a reasonable direction, considering the humans' incremental language processing.
Related to this, the measure of pseudo-perplexity might also induce unintended biases in our results, this is a common metric in NLP though.
In addition, there are different methods to fine-tune the model to multiple languages, e.g., using adapters.
Comparing these methods with our scheme will be an interesting direction.

\paragraph{Data:} There are possible variations of L1--L2 combinations.
Although we selected L1 from one of the four languages (German, French, Russian, Japanese) and fixed L2 as English, increasing the coverage of languages will lead to more generalized conclusions.
Furthermore, the performance of LMs was generally not so good on the BLiMP dataset.
We suspect that this is due to the limited L2 training data size; it is also worth exploring scaling up the experiments into typical NLP experiments.

\paragraph{Evaluation:} While our focus is on morphology, syntax, and semantic generalization, L2 acquisition studies are conducted from broader perspectives, such as the growth of vocabulary size.
In addition, our observations are  from the perspective of the LMs; the contrast between LMs' and humans' L2 learning is more important from an interdisciplinary perspective.

\subsection*{Performance was overall poor}
One reviewer is concerned that our results on the BLiMP are generally near the chance level, and it may be difficult to derive findings from such poor results.
We thank the reviewer and would like to share our thoughts and limitations here.

First, comparisons to the chance rate are not always meaningful. In BLiMP, the task is typically to select a correct generalization over an incorrect one. Occasionally, neural models overly prefer incorrect generalizations more than chance level. For instance, in the linear vs. hierarchical generalization contrast, neural models often favor linear, causing accuracy to drop near 0, far below chance~\cite{McCoy_undated-px}. In such cases, achieving accuracies around 50 indicates more than random guessing, as models avoid an excessive preference for incorrect generalizations, moving toward a more neutral stance.
Thus, we believe that it is also worthwhile to observe how much performance improves from below the chance level.

Furthermore, in BLiMP's finer-grained test suits, our models sometimes exhibit an accuracy of 0 or 100 (resulting in an overall score of around 50.0), highlighting that our models do not always act as random guessing baselines.
The full results across more fine-grained test suits are shown in Appendix~\ref{tab:full_results_l2}.

\section*{Ethics Statement}
There might be a possibility that the texts we used (CC-100 and Tatoeba) have socially biased, despite their popular use in the NLP community.
We adopted cognitively-plausible restricted settings with respect to data size, which can potentially be aligned with environmentally friendly, green NLP.

\section*{Acknowledgements}
We would like to express our gratitude for the anonymous reviewers who provided many insightful comments that have improved our paper.
Special thanks also go to the members of NAIST NLP Labolayory and Tohoku NLP Laboratory for the interesting comments and energetic discussions.
This work was supported by JSPS KAKENHI Grant Number JP19K20351.

\bibliography{anthology,custom}
\bibliographystyle{acl_natbib}

\clearpage
\appendix
\begin{table}[htbp]
    \centering
    \begin{tabular}{lr}
        \toprule
        dropout & $0.1$\\
        attention\_dropout & $0.1$\\
        accumulate\_gradients & $4$ \\
        emb\_dim & $256$ \\
        ffn\_embed\_dim & $1024$ \\
        gelu\_activation & True \\
        Optimizer &  adam\_inverse\_sqrt \\
        & lr=$0.00020$ \\
        & eps=$0.000001$ \\
        & warmup\_updates=$30000$ \\ 
        & beta1=$0.9$ \\
        & beta2=$0.999$ \\
        & weight\_decay=$0.01$ \\
        epoch &  100 \\
        n\_heads & 8 \\
        n\_layers & 12 \\
        clip\_grad\_norm & $1.0$ \\
        amp & $2$ \\
        fp16 & True \\
        \bottomrule
    \end{tabular}
\caption{Hyperparameters of the LMs}
\label{tab:hyperparameters}
\end{table}

\begin{table}[tbp]
    \centering
    \begin{tabular}{@{}lp{2.0cm}p{2.45cm}@{}}
        \toprule
        Name & Version & License \\
         \cmidrule(r){1-1} \cmidrule(lr){2-2} \cmidrule(lr){3-3}
        fastBPE & 0.1.0 & MIT License \\
        \cmidrule(r){1-1} \cmidrule(lr){2-2} \cmidrule(lr){3-3}
        kytea & 0.4.7 & Apache 2.0 \\
        \cmidrule(r){1-1} \cmidrule(lr){2-2} \cmidrule(lr){3-3}
        mosesdecoder & 0.4.0 & LGPL 2.1 \\
        \cmidrule(r){1-1} \cmidrule(lr){2-2} \cmidrule(lr){3-3}
        XLM & 0.1.0 & CC BY-NC 4.0 \\
        \cmidrule(r){1-1} \cmidrule(lr){2-2} \cmidrule(lr){3-3}
        BLiMP & 0.1.0 & CC BY-NC 4.0 \\
        \cmidrule(r){1-1} \cmidrule(lr){2-2} \cmidrule(lr){3-3}
        CC100 (CC-Net) & 1.0.0 & MIT License \\
        \cmidrule(r){1-1} \cmidrule(lr){2-2} \cmidrule(lr){3-3}
        CLAMS & 0.1.0 & Apache 2.0 \\
        \cmidrule(r){1-1} \cmidrule(lr){2-2} \cmidrule(lr){3-3}
        Tatoeba & v2022-03-03 & CC–BY 2.0 FR \\
        \bottomrule
    \end{tabular}
    \caption{The versions and licenses of used tools and datasets. These tools and datasets used in this study were designed for the purposes of research and language learning.}
    \label{tab:versions_licenses}
\end{table}

\section{Experimental Procedure}
\label{sec:appendix}
We list the hyperparameters in Table~\ref{tab:hyperparameters}. 
The versions and licenses of used tools and datasets are listed in Table~\ref{tab:versions_licenses}.

\paragraph{L1 Acquisition:}
We used mosesdecoder~\citep{mosesdecoder} as French, German and Russian tokenizer, kytea\footnote{\url{http://www.phontron.com/kytea/}}\citep{neubig-etal-kytea,neubig-mori-kytea} as Japanese tokenizer and segmented words into subwords with fastBPE\footnote{\url{https://github.com/glample/fastBPE}}\citep{sennrich-etal-2016}
The dataset was split into train/dev/test in the ratio of 8:1:1.
We set 14,000 vocabulary size for any language.
We trained our models with 4 parallel GPUs (VRAM 48G), which took 6 days per model.

\paragraph{L2 Acquisition:}
We added English tokens from the parallel corpus into BPE codes and vocabulary used in L1 Acquisition and removed duplicated tokens and vocabulary.
As for models not using a monolingual corpus, we created the codes and vocabulary using a parallel corpus of both L1 and L2.
We used mosesdecoder~\citep{mosesdecoder} as English tokenizer. As for other languages, we use tokenizers the same as L1 Acquisition.
The dataset was split into train/dev/test in the ratio of 8:1:1.
As we increased the number of vocabularies in the embedding layer, the weights/biases in the final layer were also increased.
Our four LMs were trained with different 3 seeds and reported their averages as results.
Compared models in our preliminary experiment (Sec.~\ref{sec:pre-experiment}) are shown in Figure~\ref{fig:conneau_model}.
We trained our models with 2--4 GPUs (VRAM 48G), which took around 5 hours per model.

\begin{table*}[htbp]
    \centering
    \scriptsize
    \begin{tabular}{@{}cclp{0.4cm}p{0.4cm}p{0.4cm}p{0.4cm}@{}}
        \toprule
        \multirow{2}{*}{Coarse} & \multirow{2}{*}{Specific} & \multirow{2}{*}{Challenge} & \multicolumn{4}{c}{First Language} \\
            \cmidrule(lr){4-7}
        & & & Fr & De & Ru & Ja \\
            \cmidrule(r){1-7}
        
        \multirow{18}{*}{Morphology} & \multirow{2}{*}{ANA.AGR} & anaphor\_gender\_agreement & \multicolumn{1}{r}{$61.5$} & \multicolumn{1}{r}{$23.7$} & \multicolumn{1}{r}{$52.7$} & \multicolumn{1}{r}{$57.7$} \\
         & & anaphor\_number\_agreement & \multicolumn{1}{r}{$50.1$} & \multicolumn{1}{r}{$62.5$} & \multicolumn{1}{r}{$53.1$} & \multicolumn{1}{r}{$65.4$}\\
             \cmidrule(lr){2-2} \cmidrule(lr){3-3} \cmidrule(lr){4-4} \cmidrule(lr){5-5} \cmidrule(lr){6-6} \cmidrule(lr){7-7}
         & \multirow{8}{*}{D-N AGR} & determiner\_noun\_agreement\_1 & \multicolumn{1}{r}{$80.8$} & \multicolumn{1}{r}{$74.5$} & \multicolumn{1}{r}{$69.1$} & \multicolumn{1}{r}{$67.9$} \\
         & & determiner\_noun\_agreement\_2 & \multicolumn{1}{r}{$70.1$} & \multicolumn{1}{r}{$76.5$} & \multicolumn{1}{r}{$62.6$} & \multicolumn{1}{r}{$79.8$} \\
         & & determiner\_noun\_agreement\_irregular\_1 & \multicolumn{1}{r}{$65.8$} & \multicolumn{1}{r}{$63.0$} & \multicolumn{1}{r}{$58.3$} & \multicolumn{1}{r}{$54.5$} \\
         & & determiner\_noun\_agreement\_irregular\_2 & \multicolumn{1}{r}{$64.9$} & \multicolumn{1}{r}{$72.3$} & \multicolumn{1}{r}{$56.5$} & \multicolumn{1}{r}{$78.3$} \\ 
         & & determiner\_noun\_agreement\_with\_adj\_1 & \multicolumn{1}{r}{$77.9$} & \multicolumn{1}{r}{$69.2$} & \multicolumn{1}{r}{$59.9$} & \multicolumn{1}{r}{$61.1$} \\
         & & determiner\_noun\_agreement\_with\_adj\_2 & \multicolumn{1}{r}{$62.8$} & \multicolumn{1}{r}{$66.9$} & \multicolumn{1}{r}{$56.1$} & \multicolumn{1}{r}{$65.7$} \\
         & & determiner\_noun\_agreement\_with\_adj\_irregular\_1 & \multicolumn{1}{r}{$72.7$} & \multicolumn{1}{r}{$64.1$} & \multicolumn{1}{r}{$54.0$} & \multicolumn{1}{r}{$50.5$} \\
         & & determiner\_noun\_agreement\_with\_adj\_irregular\_2 & \multicolumn{1}{r}{$61.1$} & \multicolumn{1}{r}{$63.0$} & \multicolumn{1}{r}{$51.9$} & \multicolumn{1}{r}{$68.4$} \\
             \cmidrule(lr){2-2} \cmidrule(lr){3-3} \cmidrule(lr){4-4} \cmidrule(lr){5-5} \cmidrule(lr){6-6} \cmidrule(lr){7-7}
         & \multirow{2}{*}{IRREGULAR} & irregular\_past\_participle\_adjectives & \multicolumn{1}{r}{$75.1$} & \multicolumn{1}{r}{$61.6$} & \multicolumn{1}{r}{$95.3$} & \multicolumn{1}{r}{$79.1$} \\
         & & irregular\_past\_participle\_verbs & \multicolumn{1}{r}{$70.9$} & \multicolumn{1}{r}{$77.0$} & \multicolumn{1}{r}{$50.0$} & \multicolumn{1}{r}{$62.0$} \\
             \cmidrule(lr){2-2} \cmidrule(lr){3-3} \cmidrule(lr){4-4} \cmidrule(lr){5-5} \cmidrule(lr){6-6} \cmidrule(lr){7-7}
         & \multirow{6}{*}{S-V AGR} & distractor\_agreement\_relational\_noun & \multicolumn{1}{r}{$50.9$} & \multicolumn{1}{r}{$65.5$} & \multicolumn{1}{r}{$39.8$} & \multicolumn{1}{r}{$42.6$} \\
         & & distractor\_agreement\_relative\_clause & \multicolumn{1}{r}{$48.6$} & \multicolumn{1}{r}{$51.7$} & \multicolumn{1}{r}{$46.1$} & \multicolumn{1}{r}{$45.4$} \\
         & & irregular\_plural\_subject\_verb\_agreement\_1 & \multicolumn{1}{r}{$64.2$} & \multicolumn{1}{r}{$68.9$} & \multicolumn{1}{r}{$56.7$} & \multicolumn{1}{r}{$54.4$} \\
         & & irregular\_plural\_subject\_verb\_agreement\_2 & \multicolumn{1}{r}{$65.5$} & \multicolumn{1}{r}{$74.5$} & \multicolumn{1}{r}{$63.6$} & \multicolumn{1}{r}{$58.9$} \\
         & & regular\_plural\_subject\_verb\_agreement\_1 & \multicolumn{1}{r}{$65.9$} & \multicolumn{1}{r}{$72.2$} & \multicolumn{1}{r}{$60.7$} & \multicolumn{1}{r}{$59.5$} \\
         & & regular\_plural\_subject\_verb\_agreement\_2 & \multicolumn{1}{r}{$67.3$} & \multicolumn{1}{r}{$69.2$} & \multicolumn{1}{r}{$62.5$} & \multicolumn{1}{r}{$57.2$} \\
         
         \cmidrule(r){1-7}
         
         \multirow{26}{*}{Syntax} & \multirow{9}{*}{ARG.STR} & animate\_subject\_passive & \multicolumn{1}{r}{$65.1$} & \multicolumn{1}{r}{$55.9$} & \multicolumn{1}{r}{$51.1$} & \multicolumn{1}{r}{$53.9$}\\
         & & animate\_subject\_trans & \multicolumn{1}{r}{$44.7$} & \multicolumn{1}{r}{$44.1$} & \multicolumn{1}{r}{$31.7$} & \multicolumn{1}{r}{$37.3$} \\
         & & causative & \multicolumn{1}{r}{$39.6$} & \multicolumn{1}{r}{$53.4$} & \multicolumn{1}{r}{$35.8$} & \multicolumn{1}{r}{$38.8$}\\
         & & drop\_argument & \multicolumn{1}{r}{$57.5$} & \multicolumn{1}{r}{$44.6$} & \multicolumn{1}{r}{$44.0$} & \multicolumn{1}{r}{$54.1$} \\
         & & inchoative & \multicolumn{1}{r}{$51.6$} & \multicolumn{1}{r}{$43.5$} & \multicolumn{1}{r}{$37.7$} & \multicolumn{1}{r}{$45.2$}\\
         & & intransitive & \multicolumn{1}{r}{$56.0$} & \multicolumn{1}{r}{$47.1$} & \multicolumn{1}{r}{$40.5$} & \multicolumn{1}{r}{$52.4$} \\
         & & passive\_1 & \multicolumn{1}{r}{$61.4$} & \multicolumn{1}{r}{$62.6$} & \multicolumn{1}{r}{$60.9$} & \multicolumn{1}{r}{$67.7$} \\
         & & passive\_2 & \multicolumn{1}{r}{$62.7$} & \multicolumn{1}{r}{$70.0$} & \multicolumn{1}{r}{$65.1$} & \multicolumn{1}{r}{$66.2$} \\
         & & transitive & \multicolumn{1}{r}{$59.8$} & \multicolumn{1}{r}{$56.8$} & \multicolumn{1}{r}{$56.1$} & \multicolumn{1}{r}{$53.8$} \\
            \cmidrule(lr){2-2} \cmidrule(lr){3-3} \cmidrule(lr){4-4} \cmidrule(lr){5-5} \cmidrule(lr){6-6} \cmidrule(lr){7-7}
         & \multirow{2}{*}{ELLIPSIS} & ellipsis\_n\_bar\_1 & \multicolumn{1}{r}{$52.1$} & \multicolumn{1}{r}{$50.2$} & \multicolumn{1}{r}{$46.4$} & \multicolumn{1}{r}{$46.0$}\\
         & & ellipsis\_n\_bar\_2 & \multicolumn{1}{r}{$83.2$} & \multicolumn{1}{r}{$76.8$} & \multicolumn{1}{r}{$62.1$} & \multicolumn{1}{r}{$64.5$} \\
            \cmidrule(lr){2-2} \cmidrule(lr){3-3} \cmidrule(lr){4-4} \cmidrule(lr){5-5} \cmidrule(lr){6-6} \cmidrule(lr){7-7}
         & \multirow{7}{*}{FILLER-GAP} & wh\_questions\_object\_gap & \multicolumn{1}{r}{$34.7$} & \multicolumn{1}{r}{$87.2$} & \multicolumn{1}{r}{$36.6$} & \multicolumn{1}{r}{$31.4$} \\
         & & wh\_questions\_subject\_gap & \multicolumn{1}{r}{$66.8$} & \multicolumn{1}{r}{$89.2$} & \multicolumn{1}{r}{$64.9$} & \multicolumn{1}{r}{$61.8$} \\
         & & wh\_questions\_subject\_gap\_long\_distance & \multicolumn{1}{r}{$68.8$} & \multicolumn{1}{r}{$95.5$} & \multicolumn{1}{r}{$63.8$} & \multicolumn{1}{r}{$62.4$} \\
         & & wh\_vs\_that\_no\_gap & \multicolumn{1}{r}{$68.4$} & \multicolumn{1}{r}{$97.1$} & \multicolumn{1}{r}{$60.5$} & \multicolumn{1}{r}{$63.2$} \\
         & & wh\_vs\_that\_no\_gap\_long\_distance &\multicolumn{1}{r}{$61.3$} & \multicolumn{1}{r}{$97.8$} & \multicolumn{1}{r}{$56.1$} & \multicolumn{1}{r}{$53.5$} \\
         & & wh\_vs\_that\_with\_gap & \multicolumn{1}{r}{$39.8$} & \multicolumn{1}{r}{$6.6$} & \multicolumn{1}{r}{$40.1$} & \multicolumn{1}{r}{$38.8$} \\
         & & wh\_vs\_that\_with\_gap\_long\_distance & \multicolumn{1}{r}{$42.6$} & \multicolumn{1}{r}{$4.1$} & \multicolumn{1}{r}{$45.2$} & \multicolumn{1}{r}{$47.7$} \\
            \cmidrule(lr){2-2} \cmidrule(lr){3-3} \cmidrule(lr){4-4} \cmidrule(lr){5-5} \cmidrule(lr){6-6} \cmidrule(lr){7-7}
         & \multirow{8}{*}{ISLAND} & adjunct\_island & \multicolumn{1}{r}{$49.0$} & \multicolumn{1}{r}{$52.5$} & \multicolumn{1}{r}{$50.0$} & \multicolumn{1}{r}{$56.8$ }\\
         & & complex\_NP\_island & \multicolumn{1}{r}{$47.7$} & \multicolumn{1}{r}{$43,0$} & \multicolumn{1}{r}{$43.9$} & \multicolumn{1}{r}{$53.9$} \\
         & & coordinate\_structure\_constraint\_complex\_left\_branch & \multicolumn{1}{r}{$33.7$} & \multicolumn{1}{r}{$40.2$} & \multicolumn{1}{r}{$41.1$} & \multicolumn{1}{r}{$39.0$} \\
         & & coordinate\_structure\_constraint\_object\_extraction & \multicolumn{1}{r}{$68.8$} & \multicolumn{1}{r}{$78.9$} & \multicolumn{1}{r}{$57.7$} & \multicolumn{1}{r}{$53.9$} \\
         & & left\_branch\_island\_echo\_question & \multicolumn{1}{r}{$36.6$} & \multicolumn{1}{r}{$37.2$} & \multicolumn{1}{r}{$27.1$} & \multicolumn{1}{r}{$37.2$} \\
         & & left\_branch\_island\_simple\_question & \multicolumn{1}{r}{$63.1$} & \multicolumn{1}{r}{$71.7$} & \multicolumn{1}{r}{$68.0$} & \multicolumn{1}{r}{$69.1$} \\
         & & sentential\_subject\_island & \multicolumn{1}{r}{$50.5$} & \multicolumn{1}{r}{$50.1$} & \multicolumn{1}{r}{$52.3$} & \multicolumn{1}{r}{$55.4$} \\
         & & wh\_island & \multicolumn{1}{r}{$68.0$} & \multicolumn{1}{r}{$8.1$} & \multicolumn{1}{r}{$54.5$} & \multicolumn{1}{r}{$66.6$} \\
         
        \cmidrule(r){1-7}
         
        \multirow{11}{*}{Semantics} & \multirow{7}{*}{NPI} & matrix\_question\_npi\_licensor\_present & \multicolumn{1}{r}{$46.2$} & \multicolumn{1}{r}{$46.0$} & \multicolumn{1}{r}{$14.3$} & \multicolumn{1}{r}{$30.9$} \\
        & & npi\_present\_1 & \multicolumn{1}{r}{$52.1$} & \multicolumn{1}{r}{$94.5$} & \multicolumn{1}{r}{$41.3$} & \multicolumn{1}{r}{$72.3$} \\
        & & npi\_present\_2 & \multicolumn{1}{r}{$56.4$} & \multicolumn{1}{r}{$94.8$} & \multicolumn{1}{r}{$39.5$} & \multicolumn{1}{r}{$72.5$} \\
        & & only\_npi\_licensor\_present & \multicolumn{1}{r}{$1.0$} & \multicolumn{1}{r}{$0.2$} & \multicolumn{1}{r}{$0.1$} & \multicolumn{1}{r}{$0.2$} \\
        & & only\_npi\_scope & \multicolumn{1}{r}{$35.9$} & \multicolumn{1}{r}{$55.2$} & \multicolumn{1}{r}{$50.4$} & \multicolumn{1}{r}{$52.1$} \\
        & & sentential\_negation\_npi\_licensor\_present & \multicolumn{1}{r}{$32.1$} & \multicolumn{1}{r}{$34.2$} & \multicolumn{1}{r}{$39.6$} & \multicolumn{1}{r}{$11.8$} \\
        & & sentential\_negation\_npi\_scope & \multicolumn{1}{r}{$59.6$} & \multicolumn{1}{r}{$57.6$} & \multicolumn{1}{r}{$44.6$} & \multicolumn{1}{r}{$47.1$} \\
            \cmidrule(lr){2-2} \cmidrule(lr){3-3} \cmidrule(lr){4-4} \cmidrule(lr){5-5} \cmidrule(lr){6-6} \cmidrule(lr){7-7}
         & \multirow{4}{*}{QUANTIFIERS} & existential\_there\_quantifiers\_1 & \multicolumn{1}{r}{$85.2$} & \multicolumn{1}{r}{$84.0$} & \multicolumn{1}{r}{$69.6$} & \multicolumn{1}{r}{$62.3$} \\
        & & existential\_there\_quantifiers\_2 & \multicolumn{1}{r}{$7.5$} & \multicolumn{1}{r}{$61.2$} & \multicolumn{1}{r}{$4.2$} & \multicolumn{1}{r}{$2.8$} \\
        & & superlative\_quantifiers\_1 & \multicolumn{1}{r}{$45.2$} & \multicolumn{1}{r}{$92.0$} & \multicolumn{1}{r}{$60.6$} & \multicolumn{1}{r}{$55.4$} \\
        & & superlative\_quantifiers\_2 & \multicolumn{1}{r}{$87.9$} & \multicolumn{1}{r}{$84.9$} & \multicolumn{1}{r}{$90.3$} & \multicolumn{1}{r}{$82.0$} \\

         \cmidrule(r){1-7}

         \multirow{12}{*}{Syntax \& Semantics} & \multirow{7}{*}{BINDING} & principle\_A\_c\_command & \multicolumn{1}{r}{$61.4$} & \multicolumn{1}{r}{$49.6$} & \multicolumn{1}{r}{$69.8$} & \multicolumn{1}{r}{$53.5$} \\
         & & principle\_A\_case\_1 & \multicolumn{1}{r}{$52.0$} & \multicolumn{1}{r}{$99.8$} & \multicolumn{1}{r}{$11.4$} & \multicolumn{1}{r}{$99.9$} \\
         & & principle\_A\_case\_2 & \multicolumn{1}{r}{$62.0$} & \multicolumn{1}{r}{$58.7$} & \multicolumn{1}{r}{$54.8$} & \multicolumn{1}{r}{$49.6$} \\
         & & principle\_A\_domain\_1 & \multicolumn{1}{r}{$41.3$} & \multicolumn{1}{r}{$93.7$} & \multicolumn{1}{r}{$0.7$} & \multicolumn{1}{r}{$83.9$} \\
         & & principle\_A\_domain\_2 & \multicolumn{1}{r}{$51.3$} & \multicolumn{1}{r}{$62.9$} & \multicolumn{1}{r}{$49.9$} & \multicolumn{1}{r}{$45.2$} \\
         & & principle\_A\_domain\_3 & \multicolumn{1}{r}{$52.1$} & \multicolumn{1}{r}{$49.4$} & \multicolumn{1}{r}{$48.5$} & \multicolumn{1}{r}{$48.7$} \\
         & & principle\_A\_reconstruction & \multicolumn{1}{r}{$42.4$} & \multicolumn{1}{r}{$42.2$} & \multicolumn{1}{r}{$49.6$} & \multicolumn{1}{r}{$46.3$} \\
            \cmidrule(lr){2-2} \cmidrule(lr){3-3} \cmidrule(lr){4-4} \cmidrule(lr){5-5} \cmidrule(lr){6-6} \cmidrule(lr){7-7}
          & \multirow{5}{*}{CTRL. RAIS} & existential\_there\_object\_raising & \multicolumn{1}{r}{$68.2$} & \multicolumn{1}{r}{$53.8$} & \multicolumn{1}{r}{$68.5$} & \multicolumn{1}{r}{$69.9$}\\
          & & existential\_there\_subject\_raising & \multicolumn{1}{r}{$55.1$} & \multicolumn{1}{r}{$53.3$} & \multicolumn{1}{r}{$69.7$} & \multicolumn{1}{r}{$51.1$} \\
          & & expletive\_it\_object\_raising & \multicolumn{1}{r}{$70.2$} & \multicolumn{1}{r}{$54.7$} & \multicolumn{1}{r}{$67.9$} & \multicolumn{1}{r}{$69.1$} \\
          & & tough\_vs\_raising\_1 & \multicolumn{1}{r}{$56.4$} & \multicolumn{1}{r}{$44.6$} & \multicolumn{1}{r}{$27.2$} & \multicolumn{1}{r}{$62.3$} \\
          & & tough\_vs\_raising\_2 & \multicolumn{1}{r}{$43.2$} & \multicolumn{1}{r}{$54.8$} & \multicolumn{1}{r}{$73.8$} & \multicolumn{1}{r}{$36.8$} \\
         
        \bottomrule
    \end{tabular}
    \caption{Results for each fine-grained test suit in BLiMP.}
    \label{tab:full_results_l2}
\end{table*}

\begin{table*}[t]
    \centering
    \begin{tabular}{cccccccccc}
    \toprule
         Model & L1 & Epoch & \rotatebox{90}{long\_vp\_coord} & \rotatebox{90}{obj\_rel\_within\_anim}  & \rotatebox{90}{simple\_agrmt}  & \rotatebox{90}{vp\_coord} & \rotatebox{90}{obj\_rel\_across\_anim} & \rotatebox{90}{prep\_anim} & \rotatebox{90}{subj\_rel}  \\
         \cmidrule(lr){1-1} \cmidrule(lr){2-2} \cmidrule(lr){3-3} \cmidrule(lr){4-10}
         \multirow{9}{*}{TLM-drop} & \multirow{3}{*}{Fr} & 5 & \multicolumn{1}{r}{$90.9$} & \multicolumn{1}{r}{$75.7$} & \multicolumn{1}{r}{$78.2$} & \multicolumn{1}{r}{$81.4$} & \multicolumn{1}{r}{$85.1$} & \multicolumn{1}{r}{$77.3$} & \multicolumn{1}{r}{$78.5$} \\
         & & 50 & \multicolumn{1}{r}{$75.7$} & \multicolumn{1}{r}{$66.4$} & \multicolumn{1}{r}{$71.2$} & \multicolumn{1}{r}{$72.3$} & \multicolumn{1}{r}{$69.2$} & \multicolumn{1}{r}{$69.1$} & \multicolumn{1}{r}{$70.7$} \\
         & & 100 & \multicolumn{1}{r}{$67.3$} & \multicolumn{1}{r}{$63.4$} & \multicolumn{1}{r}{$64.6$} & \multicolumn{1}{r}{$66.5$} & \multicolumn{1}{r}{$65.2$} & \multicolumn{1}{r}{$64.9$} & \multicolumn{1}{r}{$64.9$}\\
         
         \cmidrule(lr){2-2} \cmidrule(lr){3-3} \cmidrule(lr){4-10}
         & \multirow{3}{*}{De} & 5 & \multicolumn{1}{r}{$98.7$} & \multicolumn{1}{r}{$83.6$} & \multicolumn{1}{r}{$88.5$} & \multicolumn{1}{r}{$87.3$} & \multicolumn{1}{r}{$89.8$} & \multicolumn{1}{r}{$86.4$} & \multicolumn{1}{r}{$87.6$} \\
         & & 50 & \multicolumn{1}{r}{$85.9$} & \multicolumn{1}{r}{$77.4$} & \multicolumn{1}{r}{$82.9$} & \multicolumn{1}{r}{$81.3$} & \multicolumn{1}{r}{$81.3$} & \multicolumn{1}{r}{$81.0$} & \multicolumn{1}{r}{$81.4$} \\
         & & 100 & \multicolumn{1}{r}{$67.9$} & \multicolumn{1}{r}{$65.6$} & \multicolumn{1}{r}{$72.8$} & \multicolumn{1}{r}{$71.0$} & \multicolumn{1}{r}{$73.4$} & \multicolumn{1}{r}{$70.6$} & \multicolumn{1}{r}{$71.4$} \\
         
         \cmidrule(lr){2-2} \cmidrule(lr){3-3} \cmidrule(lr){4-10}
         & \multirow{3}{*}{Ru} & 5 & \multicolumn{1}{r}{$82.1$} & \multicolumn{1}{r}{$77.6$} & \multicolumn{1}{r}{$75.0$} & \multicolumn{1}{r}{$69.4$} & \multicolumn{1}{r}{$59.0$} & \multicolumn{1}{r}{$73.5$} & \multicolumn{1}{r}{$69.6$} \\
         & & 50 & \multicolumn{1}{r}{$74.1$} & \multicolumn{1}{r}{$77.0$} & \multicolumn{1}{r}{$74.7$} & \multicolumn{1}{r}{$74.6$} & \multicolumn{1}{r}{$73.5$} & \multicolumn{1}{r}{$74.1$} & \multicolumn{1}{r}{$75.3$} \\
         & & 100 & \multicolumn{1}{r}{$64.2$} & \multicolumn{1}{r}{$75.1$} & \multicolumn{1}{r}{$74.7$} & \multicolumn{1}{r}{$77.9$} & \multicolumn{1}{r}{$75.9$} & \multicolumn{1}{r}{$74.3$} & \multicolumn{1}{r}{$76.6$} \\
         
         \midrule
         \multirow{9}{*}{L1-only} & \multirow{3}{*}{Fr} & 5 & \multicolumn{1}{r}{$92.3$} & \multicolumn{1}{r}{$79.3$} & \multicolumn{1}{r}{$72.0$} & \multicolumn{1}{r}{$75.6$} & \multicolumn{1}{r}{$76.3$} & \multicolumn{1}{r}{$77.3$} & \multicolumn{1}{r}{$80.1$} \\
         & & 50 & \multicolumn{1}{r}{$97.2$} & \multicolumn{1}{r}{$82.3$} & \multicolumn{1}{r}{$74.9$} & \multicolumn{1}{r}{$77.6$} & \multicolumn{1}{r}{$78.3$} & \multicolumn{1}{r}{$77.3$} & \multicolumn{1}{r}{$80.0$}\\
         & & 100 & \multicolumn{1}{r}{$94.5$} & \multicolumn{1}{r}{$80.1$} & \multicolumn{1}{r}{$72.4$} & \multicolumn{1}{r}{$77.1$} & \multicolumn{1}{r}{$77.3$} & \multicolumn{1}{r}{$76.7$} & \multicolumn{1}{r}{$80.9$} \\

         \cmidrule(lr){2-2} \cmidrule(lr){3-3} \cmidrule(lr){4-10}
         & \multirow{3}{*}{De} & 5 & \multicolumn{1}{r}{$54.7$} & \multicolumn{1}{r}{$60.1$} & \multicolumn{1}{r}{$61.1$} & \multicolumn{1}{r}{$71.6$} & \multicolumn{1}{r}{$71.3$} & \multicolumn{1}{r}{$74.0$} & \multicolumn{1}{r}{$72.1$} \\
         & & 50 & \multicolumn{1}{r}{$56.8$} & \multicolumn{1}{r}{$58.2$} & \multicolumn{1}{r}{$55.7$} & \multicolumn{1}{r}{$67.2$} & \multicolumn{1}{r}{$67.9$} & \multicolumn{1}{r}{$71.4$} & \multicolumn{1}{r}{$70.5$} \\
         & & 100 & \multicolumn{1}{r}{$53.8$} & \multicolumn{1}{r}{$57.4$} & \multicolumn{1}{r}{$56.4$} & \multicolumn{1}{r}{$68.0$} & \multicolumn{1}{r}{$66.3$} & \multicolumn{1}{r}{$68.9$} & \multicolumn{1}{r}{$66.0$} \\

         \cmidrule(lr){2-2} \cmidrule(lr){3-3} \cmidrule(lr){4-10}
         & \multirow{3}{*}{Ru} & 5 & \multicolumn{1}{r}{$85.7$} & \multicolumn{1}{r}{$77.9$} & \multicolumn{1}{r}{$87.5$} & \multicolumn{1}{r}{$85.6$} & \multicolumn{1}{r}{$86.9$} & \multicolumn{1}{r}{$84.9$} & \multicolumn{1}{r}{$85.5$} \\
         & & 50 & \multicolumn{1}{r}{$90.8$} & \multicolumn{1}{r}{$83.2$} & \multicolumn{1}{r}{$88.6$} & \multicolumn{1}{r}{$87.7$} & \multicolumn{1}{r}{$88.1$} & \multicolumn{1}{r}{$89.3$} & \multicolumn{1}{r}{$91.8$} \\
         & & 100 & \multicolumn{1}{r}{$92.1$} & \multicolumn{1}{r}{$72.3$} & \multicolumn{1}{r}{$81.8$} & \multicolumn{1}{r}{$82.9$} & \multicolumn{1}{r}{$82.2$} & \multicolumn{1}{r}{$82.7$} & \multicolumn{1}{r}{$86.4$} \\
    \bottomrule
    \end{tabular}
    \caption{Results for each fine-grained test suit in CLAMS.}
    \label{tab:full_results_l1}
\end{table*}

\end{document}